\documentclass[hidelinks]{article}

\usepackage{PRIMEarxiv}
\usepackage{natbib}
\usepackage[utf8]{inputenc} 
\usepackage[T1]{fontenc}    
\usepackage{hyperref}       
\usepackage{url}            
\usepackage{booktabs}       
\usepackage{amsfonts}       
\usepackage{nicefrac}       
\usepackage{microtype}      
\usepackage{lipsum}
\usepackage{fancyhdr}       
\usepackage{graphicx}       
\usepackage{authblk}
\usepackage{multicol}
\usepackage{float}
\usepackage[symbol]{footmisc}
\usepackage{amssymb}
\usepackage{pdfrender}
\newcommand*{\CheckmarkBold}{%
  \textpdfrender{
    TextRenderingMode=FillStroke,
    LineWidth=.5pt, 
  }{\checkmark}%
}
\usepackage{amssymb}
\usepackage{latexsym}
\usepackage{url}
\usepackage{xcolor}
\usepackage{svg}
\usepackage{hyperref}
\usepackage{mathtools}
\usepackage{multicol}
\usepackage{multirow}
\usepackage{subfigure}
\definecolor{newcolor}{rgb}{.8,.349,.1}
\title{TFormer: 3D Tooth Segmentation in Mesh Scans with Geometry Guided Transformer}
\author[1, 2 $*$]{\textbf{Huimin Xiong}}
\author[1 $*$]{\textbf{Kunle Li}}
\author[1 $*$]{\textbf{Kaiyuan Tan}}
\author[3]{\textbf{Yang Feng}}
\author[4]{\textbf{Joey Tianyi Zhou}}
\author[5 $\dagger$]{\textbf{Jin Hao}}
\author[1, 2 $\ddagger$]{\textbf{Zuozhu Liu}}
\affil[1]{\textit{ZJU-UIUC Institute, Zhejiang University, Haining, 314400, China}}
\affil[2]{\textit{Stomatology Hospital, School of Stomatology, Zhejiang University School of Medicine, Hangzhou, 310058, China}}
\affil[3]{\textit{Angelalign Research Institute, Angel Align Inc., Shanghai, 200011, China}}
\affil[4]{\textit{A*STAR Centre for Frontier AI Research (CFAR), 138632, Singapore}}
\affil[5]{\textit{Harvard School of Dental Medicine, Harvard University, Boston, MA, USA}}
\begin{document}
\footnotetext[1]{Equal contributions}
\footnotetext[2]{Corresponding author: email: jin\_hao@g.harvard.edu}
\footnotetext[3]{Corresponding author: email: zuozhuliu@intl.zju.edu.cn}
\maketitle
\begin{abstract}
Optical Intra-oral Scanners (IOS) are widely used in digital dentistry, providing 3-Dimensional (3D) and high-resolution geometrical information of dental crowns and the gingiva. Accurate 3D tooth segmentation, which aims to precisely delineate the tooth and gingiva instances in IOS, plays a critical role in a variety of dental applications. However,
segmentation performance of previous methods are error-prone in complicated tooth-tooth or tooth-gingiva boundaries, and usually exhibit unsatisfactory results across various patients, yet the clinically applicability is not verified with large-scale dataset. 
In this paper, we propose a novel method based on 3D transformer architectures that is evaluated with large-scale and high-resolution 3D IOS datasets. Our method, termed TFormer, captures both local and  global dependencies among different teeth to distinguish various types of teeth with divergent anatomical structures and confusing boundaries. Moreover, we design a geometry guided loss based on a novel point curvature to exploit boundary geometric features, which helps refine the boundary predictions for more accurate and smooth segmentation. We further employ a  multi-task learning scheme, where an additional teeth-gingiva segmentation head is introduced to improve the performance.  Extensive experimental results in a large-scale dataset with 16,000 IOS, the largest IOS dataset to our best knowledge, demonstrate that our TFormer can surpass existing state-of-the-art baselines with a large margin, with its utility in real-world scenarios verified by a clinical applicability test. 

\end{abstract}


\begin{multicols}{2}
\section{Introduction}

Deep learning techniques have become increasingly popular in modern orthodontics treatments, e.g., automatic tooth segmentation in intra-oral scans (IOS) or cone-beam CT, panoramic X-ray segmentation, automatic treatment planning (\citet{conebeam}, \citet{panoramic}). 
Alongside the treatment process, accurate tooth segmentation in 3-Dimensional (3D) IOS dental models is the prerequisite and  a crucial component for subsequent steps, for example, multimodal fusion of IOS and CBCT, accurate tooth crown-root analysis and treatment simulation \citep{yuan2010single,kondo2004tooth}. Concretely, a 3D digitized IOS tooth model
is a mesh composed of 100,000 to 400,000 triangular faces with a spatial resolution of about 0.01mm, providing a high-fidelity digital impression of the tooth crown and the gingiva. Tooth segmentation aims at classifying each mesh face to different categories of teeth and gingiva according to the Federation Dentaire Internationale (FDI) standard \citep{herrmann1967completion}. 


There are two main streams of methods that aim to resolve this task. The conventional methods before deep learning either handle 2D images~\citep{Yamany,RangeImage,5559877} or directly operate on 3D IOS meshes~\citep{WU2014199,Yuan,Zhao,Sinthanayothin}, which usually lead to geometric information loss or rely on domain-specific knowledge from human experts with time-consuming user interactions. Recently, lots of deep learning based methods are proposed for 3D tooth segmentation in IOS. Some works train conventional convolutional neural networks or specific networks that consume simplified meshes ~\citep{TeethGNN,XJ}. In contrast, recent works, such as TSegNet (\citet{TSegNet21101949}) and DCNet (\citet{dcnet}), perform tooth segmentation by transforming the IOS meshes to point clouds, which are usually easier to process with neural networks. 

However, some key challenges remain unsolved towards an automatic and clinically applicable 3D IOS tooth segmentation solution. First, the
overall segmentation performance is still unsatisfactory on hard cases with complicated morphological topology or complex dental diseases, e.g., missing teeth, crowded teeth and erupted teeth, which frequently exist in patients seeking for orthodontic treatments (\citet{dcnet}). 
Second, current methods usually fail to precisely recognize the mesh faces between adjacent teeth or the tooth and gingiva, leading to inaccurate and jaggy boundary segmentation (\citet{dcnet}) which are problematic for downstream tasks. Finally, large-scale performance tests and clinical applicability tests are also of high necessity to fairly identify the superiority and limitations of the proposed methods, while existing works, such as MeshSegNet (\citet{meshsegnet}) and TSGCNet (\citet{zhang2021tsgcnet}), are usually evaluated with a limited amount of data samples without clinically applicability analysis. 

The aforementioned challenges motivate us to design novel methods which is able to learn expressive representations for more accurate tooth segmentation, and meanwhile, clearly delineate the boundaries with better clinical applicability in real-world applications. Inspired by the success of Transformers in various computer vision and natural language processing tasks\citet{attention,vit,han2021transformer,liu2021swin,pct,zhao2021point}, we propose a novel 3D Transformer-based framework, named TFormer, for semantic segmentation on 3D digitized IOS dental models. The IOS meshes are transformed to point clouds before being fed to TFormer. We take advantage of the self-attention mechanism to capture long-range dependencies among different teeth for effective representation learning, thus being able to cope with inherently sophisticated and inconsistent shapes and structures of the teeth.  In addition, we design a multi-task learning paradigm where another teeth-gingiva binary classification task (i.e. the auxiliary segmentation head) is introduced for acquiring finer teeth-gingiva boundaries to assist in delimiting 33 categories of teeth and gingiva. Furthermore, in view of the confusing boundary segmentation, we devise a novel geometry guided loss that is based on a newly-defined point curvature, to  perceive the implicit geometric features, which could help learn accurate, smooth and fine-grained boundaries.


We collect a large-scale, high-resolution and heterogeneous 3D IOS dataset with 16,000 complex dental models where each typically contains over 100,000 triangular faces. To the best of our knowledge, it is the largest IOS dataset to date. We also conduct extensive experiments on the dataset to demonstrate the superiority of our proposed method and verify the effectiveness of each component. Experimental results show that our model has reached 97.97\% accuracy, 94.34\% mean
intersection over union (mIoU) and 96.01\% dice similarity coefficient (DSC) on the large-scale and high-resolution IOS dataset, which exceeds the previous work by a large margin. To summarize, our main contributions are:
\begin{itemize}
\item We design a 3D Transformer-based architecture to exploit both local and global contextual information for effective representation learning, thus assisting distinguishing different types of teeth with divergent anatomical structures and confusing boundaries.
\item We design a geometry guided loss based on a novel point curvature to exploit boundary geometric features and encourage boundary refinement, leading to more accurate, smooth, and hence more clinically applicable 3D tooth segmentation. 
\item We further adopt a multi-task learning scheme, where another teeth-gingiva segmentation head is introduced to improve the segmentation performance.
\item We collect a large-scale, high-resolution and heterogeneous 3D IOS dataset for comprehensive and convincing evaluation, and conduct a clinical applicability test to examine the utility of our method in real-world scenarios. Experimental results indicate that our method has achieved state-of-the-art performance on the large-scale dataset for 3D tooth segmentation task and is much more appealing in real-world clinical applications.
\end{itemize}

The rest of the paper is organized as follows. We briefly review the related work in section \ref{sec 2}, including 3D tooth segmentation methods, 3D geometric data semantic segmentation methods and transformer for point clouds. The proposed approach is described in section \ref{sec 3}. In section \ref{sec 4}, we introduce the implementation details, comparison with competing methods, the effectiveness of different components via ablation studies and visualization results, as well as the advantages and limitations of our model. We conclude our work in section \ref{sec 5}.

\section{Related Work}
\label{sec 2}
\subsection{3D Tooth Segmentation}

 The methods for 3D intra-oral tooth segmentation fall into two categories: 1) conventional methods with deep learning, 2) deep learning based methods. As for conventional methods, some of them project the 3D mesh to 2D images, which are subsequently processed by 2D segmentation techniques and mapped back to the original mesh \cite{Yamany,RangeImage,5559877}. Other work also directly operates on meshes by extracting the curvature and topological features from meshes\cite{WU2014199,Yuan}, as well as developing interactive software and tools for tooth segmentation\cite{Zhao,Sinthanayothin}. However, these methods cannot reach applicable results and require complex feature engineering or human interactions.






Recently,  many deep learning-based approaches have been proposed for 3D tooth segmentation, either consuming the original IOS mesh data or the transformed point clouds. For example, \cite{XuCNN,Lian}  proposed a multistage convolutional neural network(CNN) and directly utilized irregular meshes. 
\cite{TeethGNN} presented a graphical neural network (GNN) TeethGNN in non-Euclidean domain for geometric feature learning. 
\cite{Sun} proposed a graphical convolutional neural network and integrated the crown shape distribution and concave boundary priors to constrain the vertex-wise labeling. 
\cite{XJ} designed TSGCNet, a two-stream GNN for separate feature extraction from mesh coordinates and normal vectors. These methods usually require time-consuming processing of the mesh data and careful architecture design to operate on IOS meshes. 


There are also a bunch of methods transforming the IOS mesh to point clouds for tooth segmentation. 
\cite{Zanjani} sampled point clouds from original meshes and designed an adversarial technique to penalize unrealistic arrangements of assigned labels.
\cite{dcnet} proposed the DCNet based on (\cite{dgcnn}) and introduced a confidence estimation module that alerts when human intervention is necessary. \cite{TSegNet21101949} also designed the TSegNet as a two-stage detection-segmentation network to deal with missing teeth. Similarly, our model takes point clouds as input. 
Unlike these previous models, we use the attention mechanism presented by \cite{attention} as our basic building block, which is naturally input order-invariant and is able to learn global representations for better segmentation. 


\subsection{Semantic Segmentation for 3D Geometric Data}

3D geometric data are most commonly represented by meshes and point clouds. Traditionally, statistical methods were used to perform segmentation on 3D data. To further improve accuracy on various segmentation tasks, deep learning based models are adopted as the major framework. There are several works for mesh segmentation. For example,  MeshCNN (\citet{meshcnn}) is a pioneering work that systematically applied a deep learning network for mesh data. 
By collapsing edges during pooling, it manages to discard unnecessary features and enhance significant ones. 
Other works like TSGCNet (\citet{TSegNet21101949}) and MeshSegNet (\citet{meshsegnet}) also perform segmentation on raw meshes. Though meshes are efficient to use for certain tasks, it is still hard to adopt them in our tooth segmentation task due to the enormous number of mesh faces, while few existing methods could directly process meshes at such a large scale. 


In addition to meshes, the point cloud is another type of 3D data that is frequently used in segmentation tasks. 
Much previous work has developed various methods to process point clouds, including point-wise, kernel-based and graph-based methods. PointNet (\citet{PointNet}) and PointNet++ (\citet{pointnet++}) are the pioneering work that developed a point-wise method to directly operate on 3D data, rather than projecting 3D data to 2D images. \citet{pat} also designed PAT to encode each point using both its absolute and relative positions to learn features using MLPs. The kernel-based method mostly designs new convolutional operators for point clouds, including PointCNN (\citet{PointCNN}), PCNN (\citet{pcnn}) and KPConv (\citet{kpconv}). DGCNN (\citet{dgcnn}) is an unprecedented work that uses a graph-based method to perform semantic segmentation on 3D data. It presented an EdgeConv module that can help capture local neighborhood information and global shape properties, without losing its permutation-invariance. Other work such as PyramNet (\citet{PyramNet}) and Graph Attention Convolution (\citet{gac}) also adopted a graph-based method to segment 3D data. These methods are proposed for the segmentation of natural objects in standard point cloud dataset, while their applicability and performance on the specific tooth point clouds need to be further enhanced. 


\begin{figure*}[!t]
\centering
\setlength{\abovecaptionskip}{0.cm}
\includegraphics[width=\linewidth]{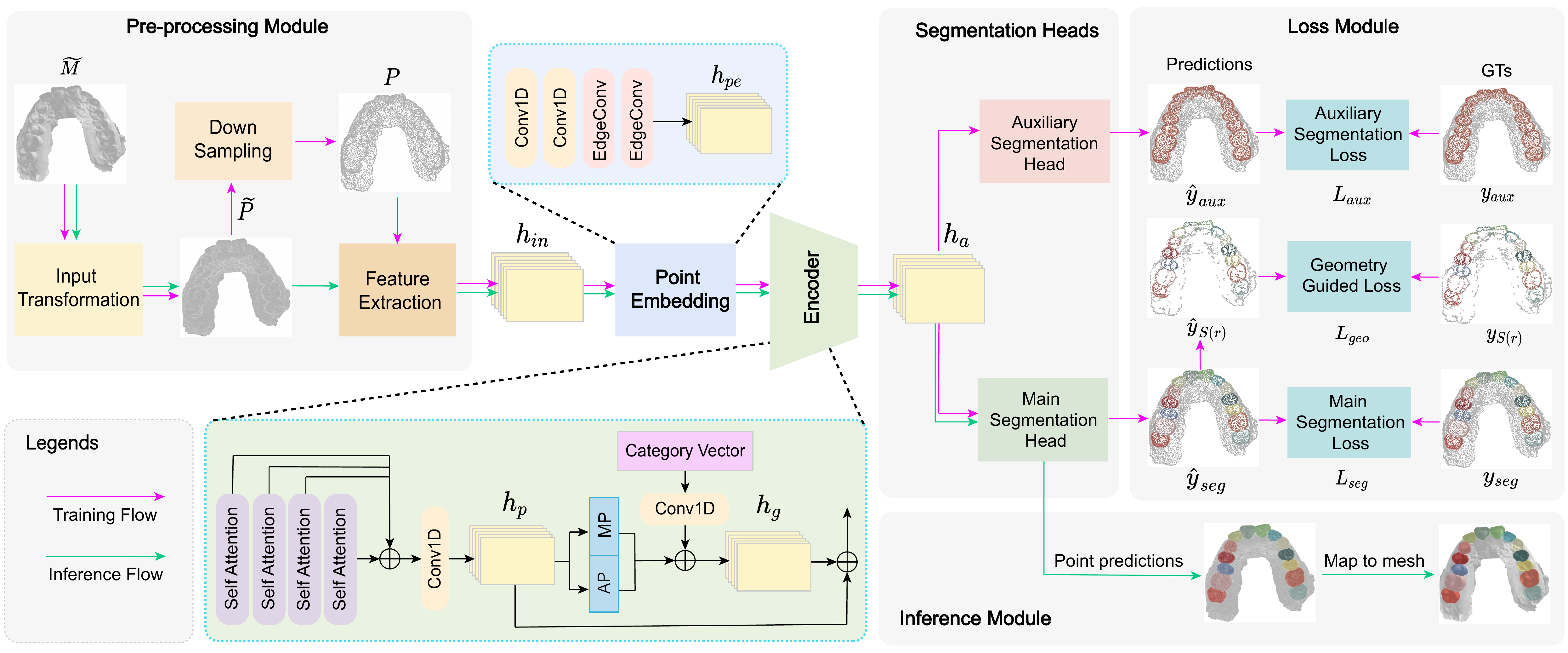}
\caption{The pipeline of our proposed TFormer for 3D tooth segmentation on digitalized IOS tooth mesh scans.}
\label{Pipeline}
\end{figure*}

\subsection{Self-Attention and Transformer}
Transformer networks (\citet{attention}) first achieved great success in Natural Language Processing (NLP), and soon got a high profile across many domains of modern deep learning. With the self-attention mechanism, Transformer models manage to perform well in holding information for long sequences, as well as get rid of the dependency on the order of input. Previous work such as BERT (\citet{bert}) and GPT-3 (\citet{gpt3}) brought pre-trained models into public view and hugely improved performance in NLP, while following work like ViT (\citet{vit}) demonstrates the effectiveness of Transformer architectures in vision.

The nature of being permutation-invariant makes Transformer models suitable to process point clouds, which consist of an unordered set of points. For example, PAT (\citet{pat}) introduced Group Shuffle Attention to replace the original multi-head attention in Transformers to outperform previous methods of classification on point clouds. Pointr (\citet{pointr}) proposed a geometric-aware block that can be plugged into any Transformer model to combine the geometric feature and the semantic feature, which helps to address the issue of lacking inductive biases of 3D structure that self-attention mechanism faces. \citet{spatialtrans} presented another perspective to apply spatial transformations to attain various local neighborhood features that can enhance the learning of the model. 
PCT (\citet{pct}) is a comprehensive work that systematically utilizes Transformer models to complete classification and segmentation tasks given 3D point clouds, and it introduced the offset-attention mechanism to replace self-attention to better handle point cloud learning.

The success of Transformer architectures in point cloud processing (e.g. \citet{zhao2021point}) has inspired us to bring it to the task of 3D tooth segmentation, though they are not directly applicable to the high-resolution IOS meshes. In our work, we propose a Transformer-based model to learn expressive representations for accurate tooth point cloud segmentation, which is greatly enhanced by task-specific architecture designs and geometry-based novel loss functions.

\section{Method}
\label{sec 3}
\subsection{Overview}
Given a 3D tooth mesh with $\widetilde{N}$ triangular faces/cells, the target of tooth segmentation is to annotate each mesh cell as one of $C=33$ semantic classes.  Mathematically, for each face $f_i$ in an IOS mesh, we need to assign it a corresponding label $l_i \in \{ 0, 11-18, 21-28, 31-38, 41-48\}$, where $0$ denotes the gingiva and the remaining denote the $32$ permanent tooth classes following the FDI notation. 

The overall pipeline of our method is illustrated in Fig. \ref{Pipeline}. 
Specifically, for each input tooth mesh $\widetilde{M}=\{{f_i}\} _{i=1}^{\widetilde{N}}$ with $\widetilde{N}$ faces, it is first converted to a tooth point cloud $\widetilde{P}=\{ {p_i} \} _{i=1}^{\widetilde{N}}$ with $\widetilde{N}$ points. Considering the high resolution of $\widetilde{P}$ (typically 100,000 to 400,000 points), we perform down-sampling on $\widetilde{P}$ to reduce the number of input points to $N$, which forms the down-sampled tooth point cloud $P=\{ {p_i} \} _{i=1}^{N}$. In the beginning, 8-dimensional features are extracted from each point $p_i$ in $P$ to get the input feature matrix $h_{in} \in \mathbb{R}^{N \times 8}$.

 During training, $h_{in}$ is fed into the point embedding module to help the model learn abundant local structure information $h_{pe}$ for each point. Thereafter, the attention-based encoder module would capture rich high-level semantic representations $h_a$ based on $h_{pe}$, which are subsequently handled by the main segmentation head and the auxiliary head together for tooth delineation. In particular, the main segmentation head will produce the prediction scores of all $N$ points over 33 classes $y_{seg} \in \mathbb{R}^{N \times 33}$, while the auxiliary head carries out the split of teeth and gingiva and generate another prediction scores $y_{aux} \in \mathbb{R}^{N \times 2}$ to assist distinguishing the tooth-gingiva boundary. 
 In this process, we devise a geometry guided loss $L_{geo}$ together with the main segmentation loss $L_{seg}$ and the auxiliary loss $L_{aux}$ to train the network to attain superior performance. 
 
 During inference, we will extract the features $h_{in} \in \mathbb{R}^{\widetilde{N} \times 8}$  for all points in $\widetilde{P}$, and generate the predictions for each point with $\lceil \frac{\widetilde{N}}{N} \rceil$ rounds of inference and map them back to the raw mesh $\widetilde{M}$. 


\subsection{Pre-process}
\subsubsection{Input Transformation and Down-sampling}
The original form of dental model is a 3D mesh which is composed of many high-resolution triangular faces. In view of the complexity of the mesh structures and the convenience of addressing 3D point clouds, we transform the raw tooth mesh $\widetilde{M}$ to a tooth point cloud $\widetilde{P}$ by taking each triangular face's gravity center. Obviously, due to the one-to-one correspondence between points in $\widetilde{P}$ and faces in $\widetilde{M}$, a tooth point cloud can also be mapped back to the respective tooth mesh on-the-fly. For the sake of enhancing the computational and memory efficiency during training, down-sampling is implemented by randomly sampling the original tooth point cloud $\widetilde{P}$ to decrease the number of points from $\widetilde{N}$ to $N$. 


\subsubsection{Feature Extraction}
Given a tooth point cloud $\widetilde{P}$/$P$, we extract 8-dimensional feature vectors $h_{in} \in \mathbb{R}^{\widetilde{N} \times 8}$/$\mathbb{R}^{N \times 8}$ for each point to preserve sufficient geometric information, including the point's 3D Cartesian coordinates, and the normal vector, Gaussian curvature and a novel point ``curvature" of the triangular face corresponding to the point. The curvatures of a triangular face are respectively computed as the sum of its three vertices' curvatures. The novel point curvature $m_i$ of a point here is defined as the mean of the angles between the point's normal vector and the normal vectors of its second-order neighbors, namely
\begin{equation}
\centering
m_i = \frac{1}{\left| K(i) \right|} \sum_{j \in K(i)} \theta(n_i, n_j), 
\label{mean curvature}
\end{equation}   
where $n_i$ denotes the $i$-th point's or central point's normal vector, $n_j$ denotes the $j$-th point's or a single adjacent point's normal vector, $K(i)$ denotes the second-order neighborhood of the $i$-th point, $\left| K(i) \right|$ represents the number of points in $K_i$, and $\theta(\cdot,\cdot)$ denotes the angle between two vectors. By definition, the edge curvature of a point reflects how much the local geometric structure around this point is curved. In other words, it presents the local geometry on 3D point clouds.

\begin{figure}[H]
    \centering
    \setlength{\abovecaptionskip}{0.cm}
    \includegraphics[width=\linewidth]{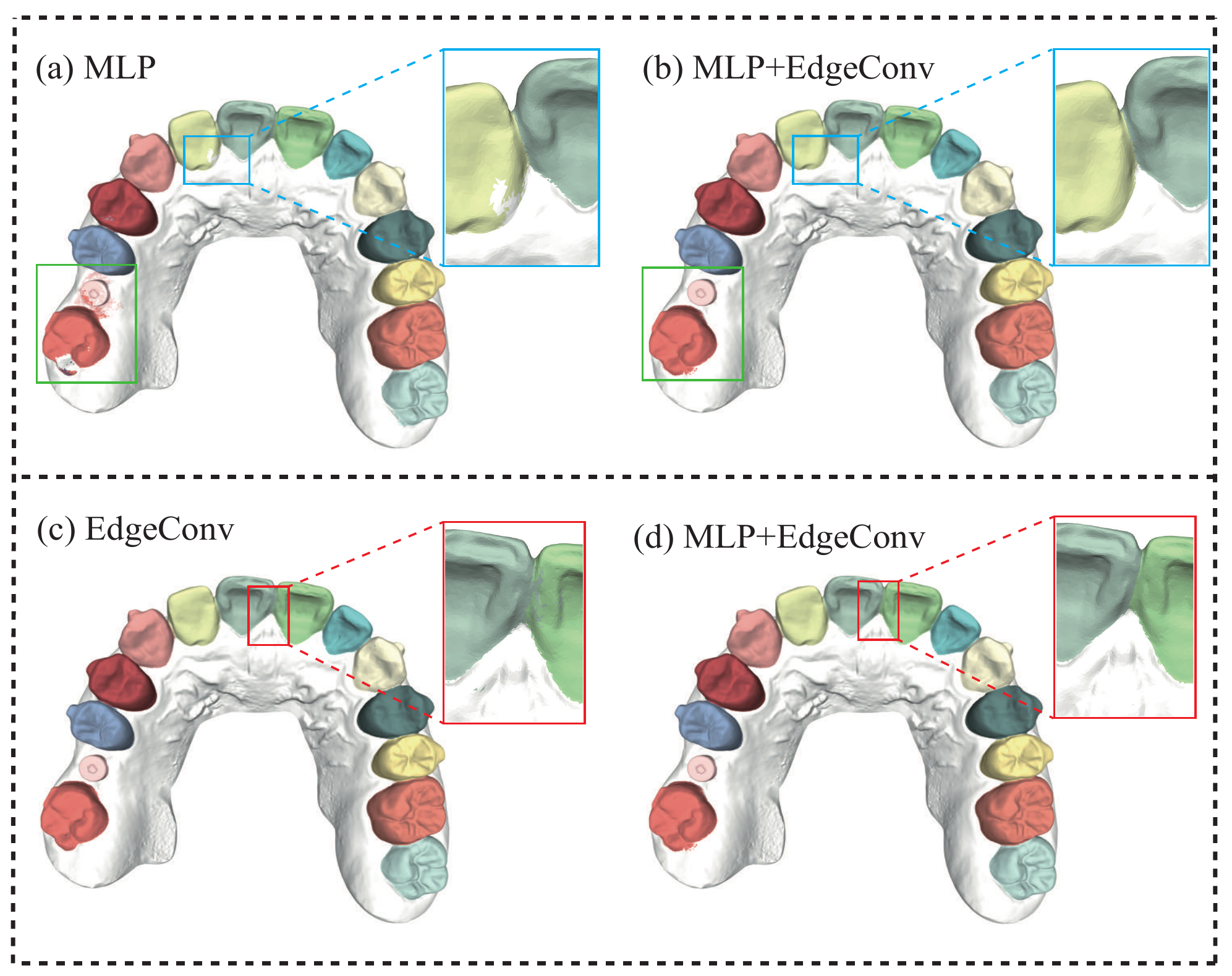}
    \caption{Segmentation of TFormer with different point embedding modules.}
    \label{pevis}
\end{figure}

\subsection{TFormer Network Architecture}

\subsubsection{Point Embedding}
The Transformer-based frameworks utilize the attention mechanism to capture long-range dependencies from a global perspective. Though global information is sufficient, the local features are usually ignored.  However, in our task, it's necessary to design an appropriate point embedding module to capture local details to complement global information for fine-grained segmentation in complex tooth structures and tooth-gingiva boundaries. 

A natural choice for point embedding is to utilize MLP to lift the initial input features to more high-level representation space for subsequent feature aggregation. However, \cite{lai2022stratified} empirically verified that MLP led to relatively slower convergence and worse performance due to lack of local geometric and contextual information. Besides, \cite{lai2022stratified} proved the importance of initial local aggregation in Transformer-based networks. Inspired by these findings, we built the point embedding module upon the \textit{EdgeConv} operation which can effectively aggregate local information \citep{dgcnn}. 


Concretely, our point embedding module is composed of two linear layers and two EdgeConv layers. For the raw input features of a tooth point cloud $h_{in}$, we have 
\begin{equation}
\centering
h_m = MLP_{pe}(h_{in}), h_m \in \mathbb{R}^{N \times d_0}.
\end{equation}
where $MLP_{pe}$ denotes the first two linear layers of the point embedding module. Here, we exploit $MLP_{pe}$ to project the input point features to high dimension space to enhance the effectiveness of succeeding local aggregation. Therefore, $MLP_{pe}$ can help identify more accurate tooth regions, as illustrated in the red squares of Fig. \ref{pevis} (c) and (d).

Thereafter, the two successive EdgeConv layers further cope with the high-level point features $h_m$, which are formulated as
\begin{equation}
\centering  
h_{{e}_{i}}^{1} = \mathop{max}\limits_{j\in \Omega(i)} ReLU \{\theta_1 \cdot (h_{m_{j}} - h_{m_{i}})+ \varphi_1 \cdot h_{m_{i}}\},
\label{EdgeConv op1}
\end{equation}
\begin{equation}
\centering  
h_{e_{i}}^2 = \mathop{max}\limits_{j\in \Omega(i)} ReLU \{\theta_2 \cdot (h_{{e}_{j}}^1 - h_{e_{i}}^{1})+ \varphi_2 \cdot h_{e_{i}}^1\},
\label{EdgeConv op2}
\end{equation}
where $h_{m_{i}} \in \mathbb{R}^{d_0}$ denotes the $i$-th feature in $h_m$, $\Omega(i)$ denotes the nearest neighborhood of the $i$-th point; $h_{{e}_{i}}^{t}$ represents the $i$-th feature vector of the $t$-th EdgeConv layer's output, $\theta_t$ and $\varphi_t$ are learnable parameters which are implemented as shared MLPs in our framework, and $i=1,2,...,N$, and $t=1,2$. 
With Equation \ref{EdgeConv op1} and \ref{EdgeConv op2}, the EdgeConv operation is repeated twice for each point $p_i$ in a static manner, and the receptive field is enlarged simultaneously. ``Static" means that the $k$ nearest neighbor graph for the same point remains fixed in two EdgeConv layers, rather than being dynamically updated alongside the depth of EdgeConv layers as in \cite{dgcnn}. The output of point embedding is $h_{pe}= \{ h_{e_{i}}^2 \} _{i=1}^{N} \in \mathbb{R}^{N \times d_{e}}$. 

With the point embedding module, the local topological information that is inherently lacking in the point clouds can be recovered to some extent, thus enriching the representation power. In other words, a point is able to interact with the points around it, where the point embodies the structural features of a local patch rather than merely comprises the raw information of its own geometric features. As shown in the green squares and the blue squares of Fig. \ref{pevis} (a) and (b), the addition of EdgeConv spots more complete local details.

\begin{figure}[H]
\centering
\setlength{\abovecaptionskip}{0.cm}
\includegraphics[width=\linewidth]{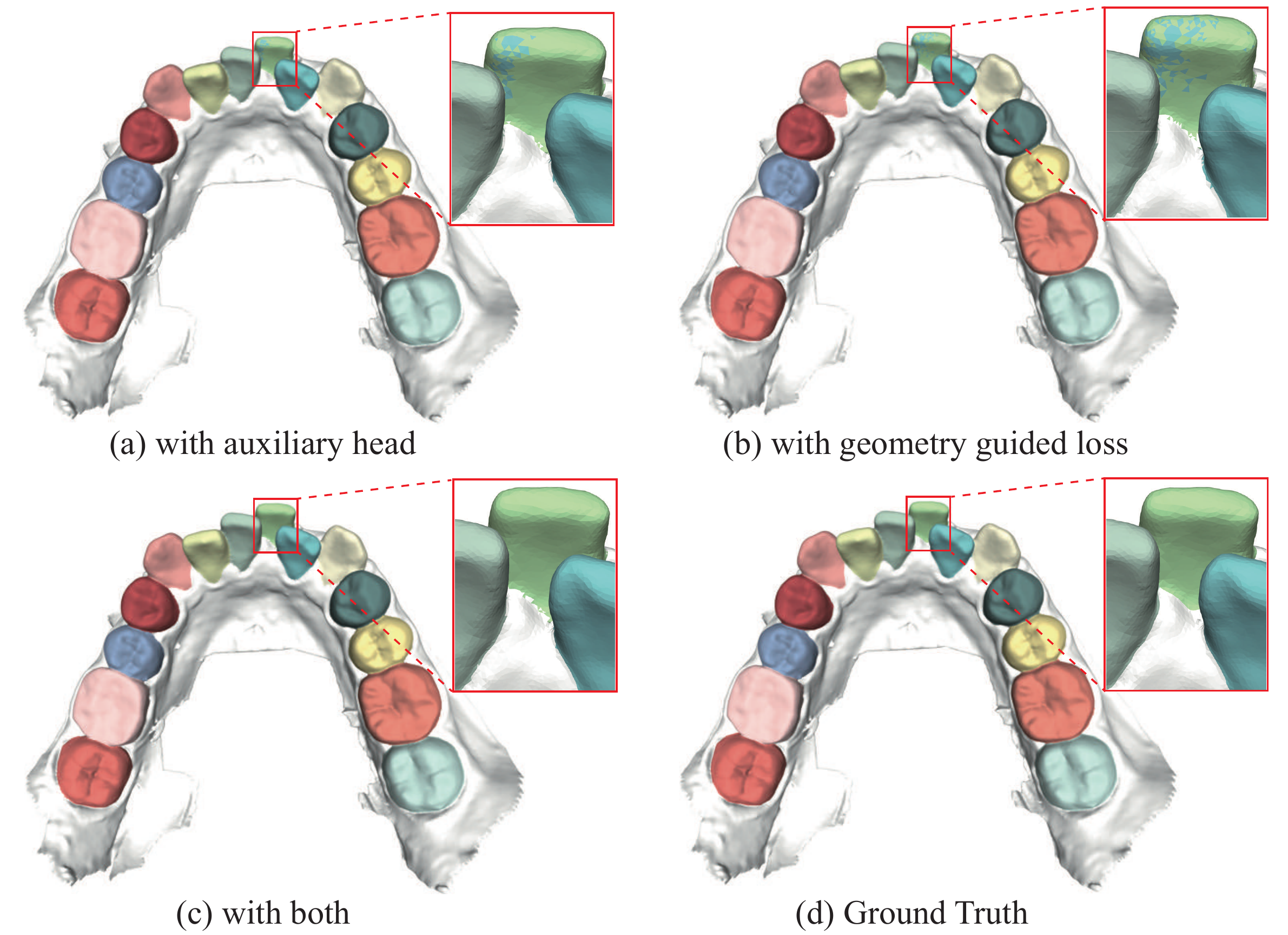}
\caption{Segmentation of TFormer with different main components.}
\label{gingiva loss vis}
\end{figure}

\subsubsection{Encoder}
The encoder module further learns the semantic representations of tooth point clouds from the output of point embedding module $h_{pe}$. First, $h_{pe}$ is tackled by four successive self-attention layers, in which the weights of query and key are shared to decrease the memory consumption and computational burden yet with satisfactory experimental results. Assume $h_{sa_0} = h_{pe}$, the self-attention part can be formulated as
\begin{equation}
\centering
h_{sa_i} = SA_i(h_{sa_{i-1}}) + h_{sa_{i-1}}, i=1,2,3,4.
\end{equation}
where $SA_i$ denotes the $i$-th self attention layer and $h_{sa_i}$ is the corresponding output. In this process, the residual connection is adopted as well. $SA_i$ is calculated as
\begin{equation}
\centering
SA_i = softmax(\frac{Q_i \cdot K_i^T}{\sqrt{d_k}}) \cdot V_i,
\end{equation}
where $Q_i \in \mathbb{R}^{N \times d_q}$, $K_i \in \mathbb{R}^{N \times d_k}$ and $V_i \in \mathbb{R}^{N \times d_v}$ are the \textit{query}, \textit{key} and \textit{value} matrices, respectively \cite{attention}. Normally, we have $d_q=d_k$. The computation is defined as 
\begin{equation}
\centering
(Q_i, K_i, V_i) = h_{sa_{i-1}} \times (W_{q_i}, W_{k_i}, W_{v_i}),
\end{equation}
where $\times$ denotes matrix multiplication and $W_{q_i}, W_{k_i}, W_{v_i}$ are different linear transformation parameters. In our implementation, we have $W_{q_i}=W_{k_i} \neq W_{v_i}$, as mentioned above.

The output features from the four attention layers are concatenated along the channel dimension, after which a linear transformation is applied to form the point feature maps $h_p$, as expressed as
\begin{equation}
h_{p} = \sigma \left( h_{sa_1} \oplus h_{sa_2} \oplus h_{sa_3} \oplus h_{sa_4} \right), 
\end{equation} 
where  $\sigma(\cdot)$ is implemented as a linear layer, and $\oplus$ denotes the channel-wise concatenation.

In order to avoid the misjudgment of jaw categories, a 2-dimensional category vector $V$ is taken as the input of our network additionally to help distinguish the maxillary and mandible. The category vector is lifted to high-dimensional feature space by a linear layer as well. Hereafter, we combine the transformed category vector with the max pooling and average pooling results of point feature maps $h_p$ to get the global feature maps $h_g$, namely 
\begin{equation}
h_g = \sigma(V) \oplus MP(h_p) \oplus AP(h_p),
\end{equation}
where ``MP" and ``AP" denote the max pooling and average pooling, respectively.

Finally, by concatenating the point feature $h_p$ and global feature $h_g$, we obtain the feature maps $h_a$ for all points, i.e.
\begin{equation}
h_a = h_p \oplus h_g.
\end{equation}

\begin{figure}[H]
\centering
\subfigure[Front view]{
\includegraphics[width=0.45\linewidth]{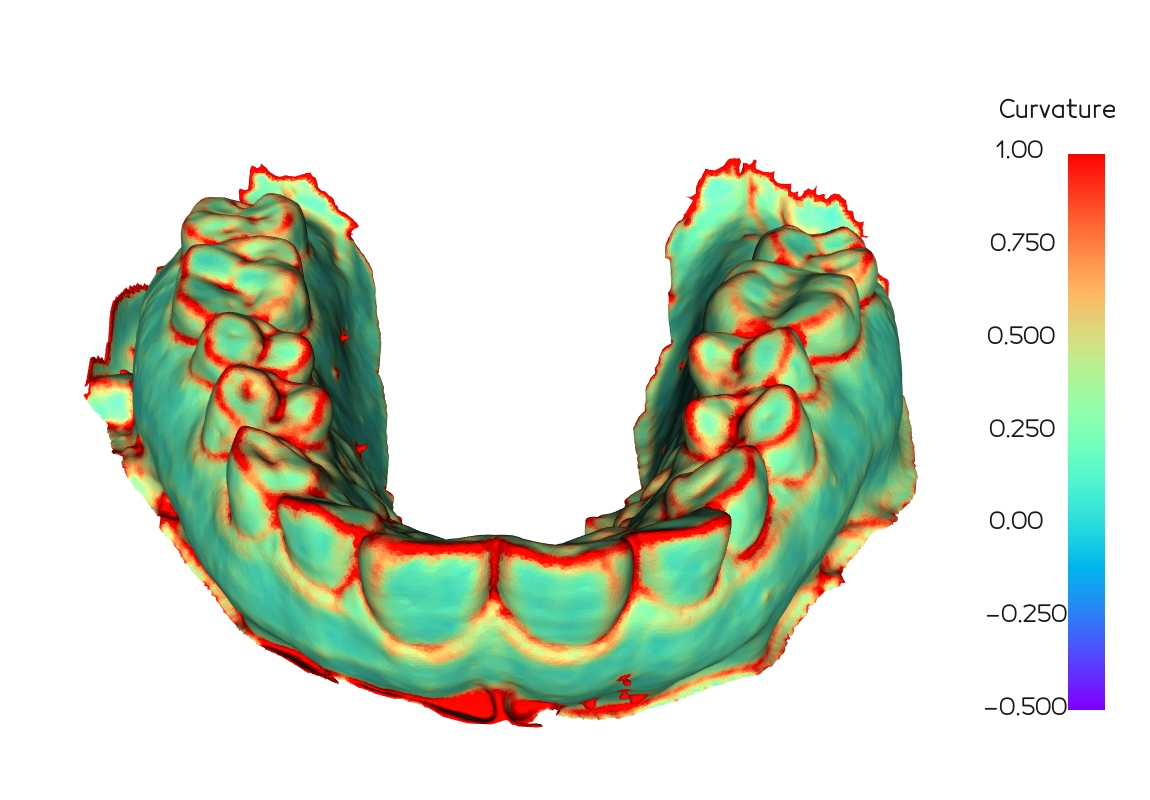}
}
\quad
\subfigure[Vertical view]{
\includegraphics[width=0.45\linewidth]{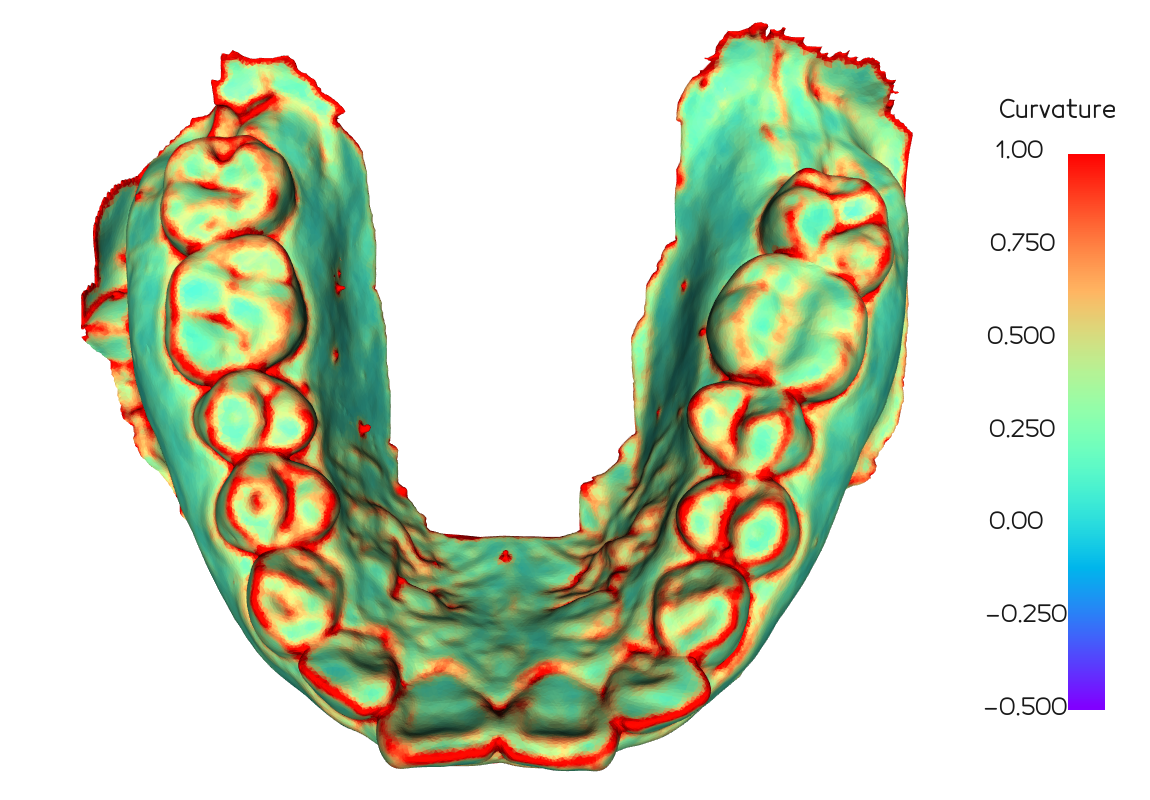}
}
\quad
\subfigure[End view (left)]{
\includegraphics[width=0.45\linewidth]{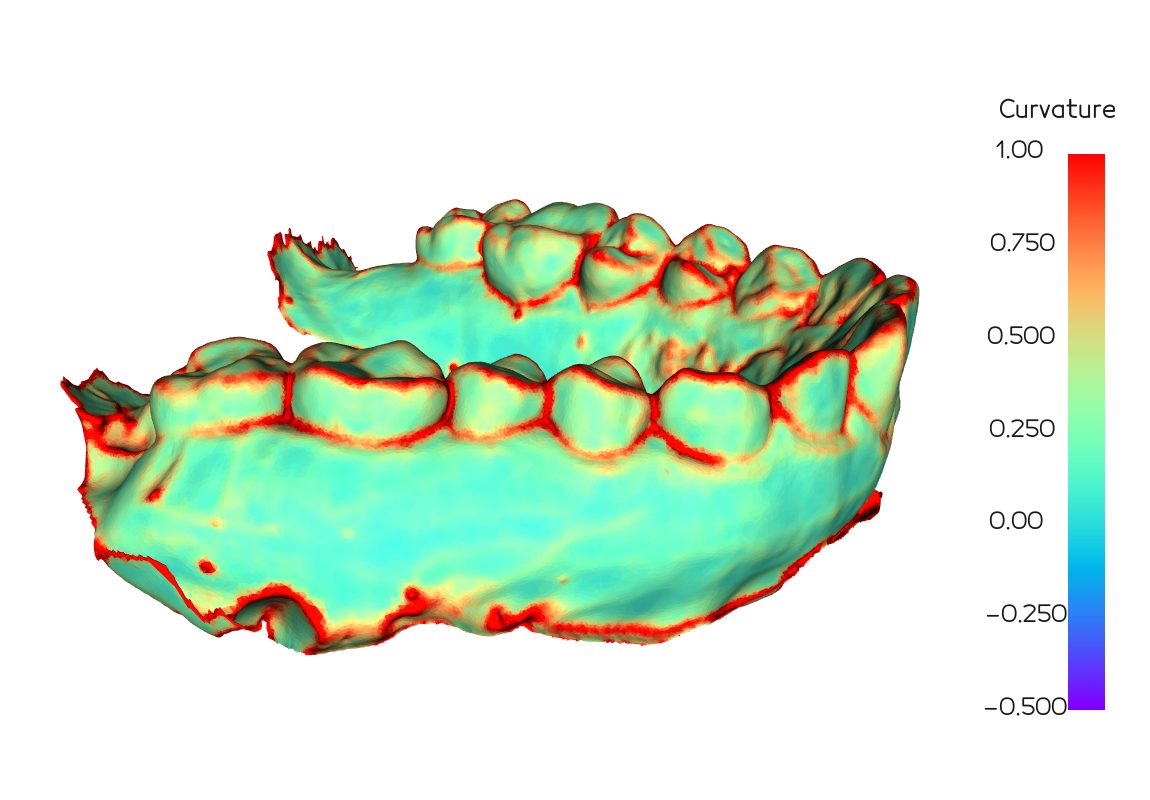}
}
\quad
\subfigure[End view (right)]{
\includegraphics[width=0.45\linewidth]{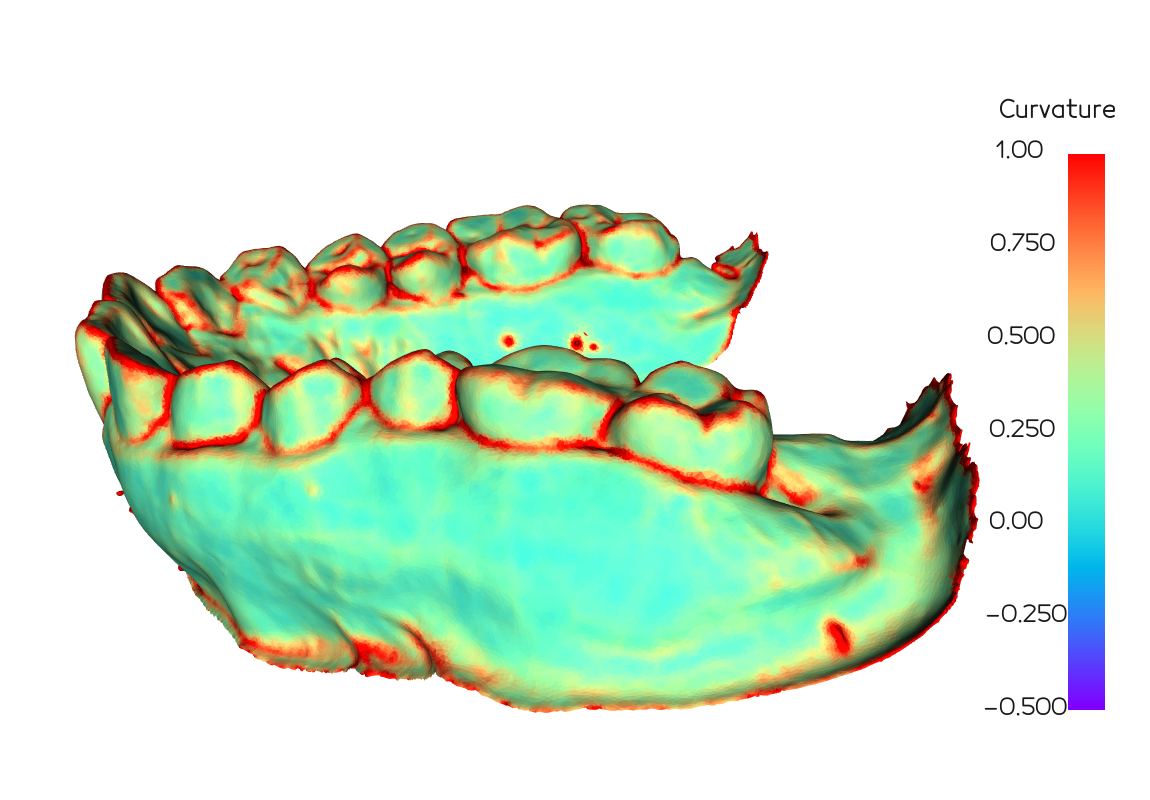}
}
\caption{Visualization of point curvatures in a point cloud.}
\label{vis cur}
\end{figure}

\subsubsection{Segmentation Heads}
\label{segmentation heads}
For the sake of enhancing the recognition capability of the network for different tooth and the gingiva categories, we design two novel segmentation heads. The main segmentation head, which is simply implemented as a MLP($MLP_{seg}$), deals with $h_a$ to yield the classification score of each point belonging to 33 classes $\hat{y}_{seg} \in \mathbb{R}^{N \times 33}$, i.e. 
\begin{equation}
\hat{y}_{seg} = MLP_{seg}(h_a).
\end{equation}

Considering the prevalence of incorrect prediction of teeth and the gingiva, we take an auxiliary segmentation head for binary classification of teeth and gum, which essentially focuses on the boundary between them. Similarly, the auxiliary segmentation head is implemented as another MLP($MLP_{aux}$) and outputs the classification scores of each point belonging to teeth and gingiva classes $\hat{y}_{aux} \in \mathbb{R}^{N \times 2}$, i.e. 
\begin{equation}
\hat{y}_{aux} = MLP_{aux}(h_a).
\end{equation}
The experimental results indicate that the cooperation with $MLP_{aux}$ can not only refine the tooth-gingiva segmentation boundary but also correct some isolated class prediction errors.
As shown in Fig. \ref{gingiva loss vis} (b) and (c), in the complex case with irregular tooth arrangement, the auxiliary segmentation head assists correcting the false predictions of the incisor greatly.

\begin{figure}[H]
\centering
\setlength{\abovecaptionskip}{0.cm}
\includegraphics[width=\linewidth]{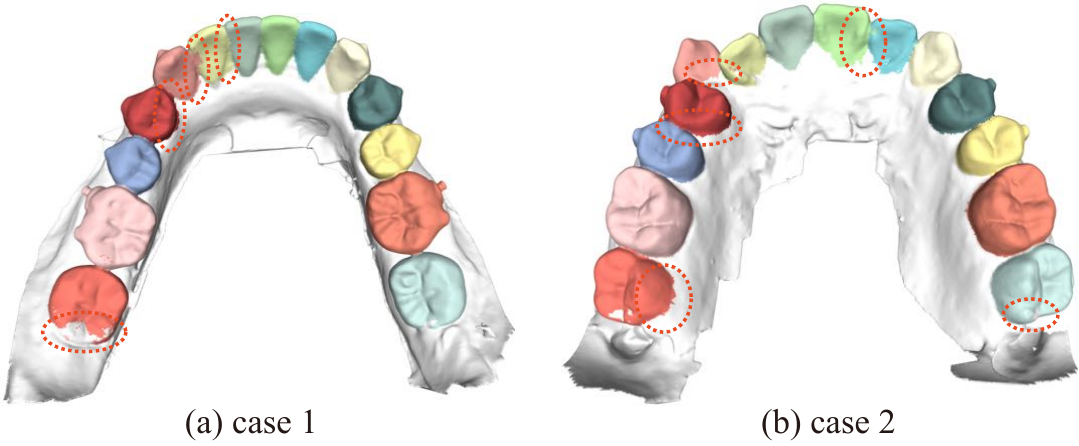}
\caption{Error-prone predictions in tooth-tooth and tooth-gingiva boundaries that commonly exist in previous methods (MeshSegNet~\citep{meshsegnet}).}
\label{boundary issues}
\end{figure}


\subsubsection{Geometry Guided Loss}
Previous methods are usually unsatisfactory to delineate the complicated tooth-tooth and tooth-gingiva boundaries across various patients. In the TFormer, we devise the geometry guided loss tailored for the tooth segmentation task, which focuses on the critical error-prone regions in a geometry-aware manner while incurring negligible extra
computations. We are inspired by such a fact that points with high point curvatures often correspond to the boundaries between teeth and gingiva or adjacent teeth and the upper sharp ends of tooth crowns (as shown in Fig. \ref{vis cur}), where the boundaries are usually the areas prone to mispredictions (as illustrated in Fig.  \ref{boundary issues}). 

On the basis of the foregoing, we define the geometry guided loss that encourage the TFormer to adaptively focus more on error-prone boundary points with a higher point curvatures. Specifically, it is defined as 
\begin{equation}
\centering
L_{geo} = - \sum _{i \in {S(r)}} \sum_{c=1}^{33}
(1-{\hat{p}}_{ic}^{geo})^{\gamma} \cdot \Phi(y_{S(r)_i}, c) \cdot log(\hat{p}_{ic}^{geo}),
\end{equation}
where $\gamma$ is the modulating factor, which is empirically set to 2 in our experiments; $y_{S(r)_i} \in \mathbb{R}^{33}$ represents the gold label of the $i$-{th} point in the point set $S(r)$; $\hat{p}_{ic}^{geo}$ denotes the predicted probability of the $i$-{th} point belonging to the $c$-{th} class, and $\Phi(\cdot, \cdot)$ is an indicator function with the form of
\begin{equation}
\Phi(y_i, c)=\left\{
\begin{aligned}
1 & , & if \; y_i=c, \\
0 & , & otherwise.
\end{aligned}
\right.
\end{equation}
Concretely, $S(r)$ is the set of error-prone points with high point curvatures, i.e. the points with top $r\space(0<r \leq 1)$ point curvatures $m_i$ defined in Equation \ref{mean curvature}. It is defined as
\begin{equation}
\underbrace{m_{a_1} \geq m_{a_2} \geq \cdots > m_{a_{rN}}}_{S(r):=\{a_1,\ a_2,\ \cdots,\ \lceil a_{rN} \rceil \}} \geq m_{a_{rN+1}} \geq \cdots \geq m_{a_N},
\end{equation}
where $a_i, i=1, 2, \cdots, N$ denotes the points sorted with a decreasing order of point curvature. We denote the prediction scores of these points as $\hat{y}_{S(r)}$, as shown in Fig. \ref{Pipeline}. 



The geometry guided loss can encourage our network to catch the hard boundary areas which typically have lower confidence than the upper sharp ends of tooth crowns, and gives these hard areas higher specific gravity, similar to the focal loss~\citep{lin2017focal}. 
Hence, the false predictions in boundary areas can be corrected effectively, enhancing the generalization ability of the network on unseen cases. As shown in the complicated case in Fig. \ref{gingiva loss vis}, 
the model with geometry guided loss (Fig. \ref{gingiva loss vis}.(c)) achieves almost perfect recognition for the incisor, while the model without geometry guided loss (Fig. \ref{gingiva loss vis}.(a)) still misclassify part of the incisor as the adjacent lateral incisor.

Moreover, the normal cross entropy loss is employed as the loss of main segmentation head and auxiliary segmentation head as well, which are respectively denoted as $L_{seg}$ and $L_{aux}$:
\begin{equation}
L_{seg} = - \sum _{i=1}^{N} \sum_{c=1}^{33} 
\Phi(y_{{seg}_i}, c) \cdot log(\hat{p}_{ic}^{seg}),
\end{equation}
\begin{equation}
L_{aux} = - \sum _{i=1}^{N}  y_{{aux}_i} \cdot log(\hat{p}_{i}^{aux}) + (1-y_{{aux}_i}) \cdot log(1 - \hat{p}_{i}^{aux}),
\end{equation}
where $\hat{p}_{ic}^{seg}$ denotes the predicted probability of the $i$-th point belonging to the $c$-th class from the main segmentation head, $\hat{p}_{i}^{aux}$ denotes the predicted probability of the $i$-th point belonging to the gingiva class from the auxiliary head, and $y_{{seg}_i} \in \mathbb{R}^{33}$ and $y_{{aux}_i} \in \mathbb{R}^{2}$ represent the ground truth (GT) labels of the $i$-th point in the main and auxiliary segmentation heads.

The total loss $L_{total}$ is computed by combining the two cross entropy losses $L_{seg}$, $L_{aux}$ for \textit{all points} and the geometry guided loss $L_{geo}$ for \textit{hard points} in the tooth point cloud:
\begin{equation}
 L_{total} = L_{seg} + \omega_{geo} \cdot L_{geo} + \omega_{aux} \cdot L_{aux},
\end{equation}
where $\omega_{geo}$ and $\omega_{aux}$ are corresponding coefficients. 



\begin{table*}[htb]
\centering
\setlength{\abovecaptionskip}{0pt}
\setlength{\belowcaptionskip}{8pt}
\caption{Main segmentation results (Tested on 1,000 patients).}
\label{comparison with baseline}
\centering
\resizebox{\textwidth}{!}{%
\begin{tabular}{c|ccc|ccc|ccc}
\hline
\multirow{2}{*}{Method} &
  \multicolumn{3}{c|}{Mandible} &
  \multicolumn{3}{c|}{Maxillary} &
  \multicolumn{3}{c}{All} \\ \cline{2-10} 
            &  mIoU  &  DSC   &  Acc   &  mIoU  &  DSC   &  Acc   &  mIoU  &  DSC  &  Acc  \\
            & (95\% CI)  & (95\% CI)   & (95\% CI)   & (95\% CI)  & (95\% CI)   & (95\% CI)   & (95\% CI)  & (95\% CI)  & (95\% CI)  \\\hline\hline
PointNet++  & 81.11 & 85.33 & 94.96 & 83.89 & 87.12 & 96.28 & 82.57 & 86.27 & 95.65\\
            &  (80.51, 81.72)   &  (84.76, 85.89)   &  (94.69, 95.22)   &  (83.38, 84.39)   &  (86.64, 87.59)   &  (96.07, 96.48)   &  (82.18, 82.97)   &  (85.90, 86.63)   &  (95.48, 95.82) \\ 
DGCNN       & 92.41 & 94.49 & 97.68 & 93.82 & 95.61 & 98.01 & 93.15 & 95.08 & 97.85 \\
            &  (91.97, 92.85)   &  (94.07, 94.91)   &  (97.50, 97.86)   &  (93.44, 94.20)   &  (95.25, 95.97)   &  (97.87, 98.15)   &  (92.86, 93.44)   &  (94.80, 95.35)   &  (97.74, 97.97) \\ \hline 
point transformer  & 92.61 & 94.83 & 97.55 & 93.93 & 95.72 & 98.06 & 93.3 & 95.3 & 97.81 \\
            &  (92.14, 93.09)   &  (94.37, 95.28)   &  (97.32, 97.77)   &  (93.56, 94.30)   &  (95.37, 96.07)   &  (97.92, 98.19)   &  (93.01, 93.60)   &  (95.01, 95.58)   &  (97.69, 97.94) \\ 
PVT         & 90.66 & 93.59 & 96.64 & 92.46 & 94.72 & 97.44 & 91.6 & 94.19 & 97.06 \\
            &  (90.16, 91.15)   &  (93.14, 94.04)   &  (96.44, 96.84)   &  (92.05, 92.87)   &  (94.34, 95.11)   &  (97.29, 97.59)   &  (91.28, 91.92)   &  (93.89, 94.48)  &  (96.93, 97.18) \\ \hline 
MeshSegNet  & 79.52 & 85.06 & 90.67 & 83.15 & 87.89 & 92.66 & 81.43 & 86.54 & 91.72 \\
            &  (78.78, 80.27)   &  (84.35, 85.76)   &  (90.26, 91.08)   &  (82.53, 83.77)   &  (87.31, 88.47)   &  (92.38, 92.94)   &  (80.94, 81.91)   &  (86.09, 87.00)    &  (91.47, 91.96) \\ 
TSGCNet     & 80.71 & 85.23 & 92.78 & 80.97 & 85.28 & 93.86 & 80.85 & 85.25 & 93.34 \\
           &   (79.90, 81.52)  &   (84.48, 85.98)   &   (92.38, 93.17)   &   (80.18, 81.77)   &   (84.55, 86.01)  &   (93.49, 94.22)  &   (80.28, 81.42)  &   (84.73, 85.78)   &   (93.07, 93.61) \\ 
DCNet       & 91.18 & 93.89 & 97.11 & 92.78 & 95.18 & 97.44 & 92.02 & 94.57 & 97.28 \\
            &  (90.76, 91.59)   &  (93.49, 94.29)   &  (96.94, 97.27)   &  (92.41, 93.14)   &  (94.83, 95.53)   &  (97.30, 97.57)   &  (91.74, 92.29)   &  (94.31, 94.84)   &  (97.17, 97.39) \\ \hline\hline 
TFormer (Our)     & \textbf{93.53}  & \textbf{95.36}  & \textbf{97.72}  & \textbf{95.07}  & \textbf{96.60}  & \textbf{98.20}  & \textbf{94.34}  & \textbf{96.01}   & \textbf{97.97} \\
            &  (93.04, 94.02)   &  (94.88, 95.84)   &  (97.47, 97.97)   &  (94.70, 95.44)   &  (96.25, 96.96)   &  (98.05, 98.35)   &  (94.03, 94.64)   &  (95.72, 96.31)   &  (97.83, 98.11) \\ \hline
\end{tabular}%
}
\end{table*}

\begin{table*}[htb]
\centering
\setlength{\abovecaptionskip}{0pt}
\setlength{\belowcaptionskip}{8pt}
\caption{Clinical applicability test on the external IOS dataset with 200 cases. 
``\#success'' and ``\#fail'' denotes the number of segmentation that meets or does not meet the clinical criteria. \#param: number of parameters in the neural network. Inf-T: end-to-end inference time for a single IOS. }
\label{clinical analysis}

\begin{tabular}{c|c|c|c|c|c}
\hline
  Model &
  \multicolumn{1}{c|}{\#success} &
  \multicolumn{1}{c|}{\#fail} &
  \multicolumn{1}{c|}{clinical error rate (\%)} &
  \multicolumn{1}{c|}{\#param} &
  \multicolumn{1}{c}{Inf-T(s)} \\ \hline\hline
\multicolumn{1}{c|}{MeshSegNet}        & 1   & 199 & 99.5 & 1.81M & 182.79 \\
\multicolumn{1}{c|}{TSGCNet}           & 15  & 185 & 92.5 & 4.13M & 31.40  \\
\multicolumn{1}{c|}{Point Transformer} & 97  & 103 & 51.5 & 6.56M & 437.21 \\
\multicolumn{1}{c|}{DCNet }            & 109 & 91  & 45.5 & \textbf{1.70M} & \textbf{5.79}   \\ \hline
\multicolumn{1}{c|}{TFormer (Our)}     & \textbf{152} & \textbf{48}  & \textbf{24.0}   & 4.21M & 23.15 \\ \hline
\end{tabular}%
\end{table*}

\begin{table*}[htb]
\centering
\setlength{\abovecaptionskip}{0pt}
\setlength{\belowcaptionskip}{8pt}
\caption{Ablation study on different components.}
\resizebox{0.8\textwidth}{!}{%
\begin{tabular}{cc|ccc|ccc|ccc}
\hline
\multicolumn{2}{c|}{Component} & \multicolumn{3}{c|}{Mandible} & \multicolumn{3}{c|}{Maxillary} & \multicolumn{3}{c}{All} \\ \hline
Geometry guided loss&  Auxiliary branch & mIoU & DSC  & Acc   & mIoU & DSC  & Acc   & mIoU & DSC  & Acc \\ \hline\hline
                    &                   &92.19 &94.27 &97.35  &94.02 &95.69 &97.96  &93.15 &95.01 &97.67  \\
                    &    \CheckmarkBold &92.53 &94.57 &97.43  &94.45 &96.10 &98.02  &93.54 &95.37 &97.74 \\
      \CheckmarkBold&                   &92.46 &94.47 &97.37  &94.45 &96.12 &98.02  &93.51 &95.33 &97.71  \\
      \CheckmarkBold&    \CheckmarkBold &\textbf{92.95} &\textbf{94.88} &\textbf{97.57}  &\textbf{94.46} &\textbf{96.07} &\textbf{98.14}  &\textbf{93.77} &\textbf{95.51} &\textbf{97.87}\\ \hline
\end{tabular}%
}
\label{ablation-geo&aux}
\end{table*}

\begin{table*}[htb]
\centering
\setlength{\abovecaptionskip}{0pt}
\setlength{\belowcaptionskip}{8pt}
\caption{Segmentation performance of the geometry guided loss with DCNet.}
\label{DCNet+loss}
\resizebox{0.7\textwidth}{!}{%
\begin{tabular}{c|ccc|ccc|ccc}
\hline
\multirow{2}{*}{Model} & \multicolumn{3}{c|}{Mandible} & \multicolumn{3}{c|}{Maxillary} & \multicolumn{3}{c}{All} \\ \cline{2-10} 
                & mIoU & DSC  & Acc  & mIoU & DSC  & Acc  & mIoU & DSC  & Acc  \\ \hline\hline
DCNet           &87.72 &91.00 &95.99 &90.77 &93.50 &96.87 &89.32 &92.31 &96.46   \\
DCNet+$L_{geo}$ &\textbf{89.66} &\textbf{92.58} &\textbf{96.46} &\textbf{91.75} &\textbf{94.28} &\textbf{97.14} &\textbf{90.75} &\textbf{93.47} &\textbf{96.82} \\ \hline
\end{tabular}%
}
\end{table*}

\begin{table*}[htb]
\centering
\setlength{\abovecaptionskip}{0pt}
\setlength{\belowcaptionskip}{8pt}
\caption{Segmentation performance of TFormer under different point embedding strategies.}
\label{point-embed}
\resizebox{0.8\textwidth}{!}{%
\begin{tabular}{cc|ccc|ccc|ccc}
\hline
\multicolumn{2}{c|}{Point embedding strategy} & \multicolumn{3}{c|}{Mandible} & \multicolumn{3}{c|}{Maxillary} & \multicolumn{3}{c}{All} \\ \hline
    MLP-based &      EdgeConv & mIoU & DSC  & Acc  & mIoU & DSC  & Acc  & mIoU & DSC  & Acc  \\ \hline \hline 
              & \CheckmarkBold&92.34 &94.42 &97.41 &94.19 &95.86 &98.00 &93.32 &95.18 &97.72   \\
\CheckmarkBold&               &89.89 &92.66 &96.52 &92.53 &94.62 &97.54 &91.28 &93.69 &97.05   \\
\CheckmarkBold& \CheckmarkBold&\textbf{92.95} &\textbf{94.88} &\textbf{97.57}  &\textbf{94.46} &\textbf{96.07} &\textbf{98.14} &\textbf{93.77} &\textbf{95.51} &\textbf{97.87} \\ \hline
\end{tabular}%
}
\end{table*}


\begin{figure*}[!t]
\centering
\setlength{\abovecaptionskip}{0.cm}
\subfigure[Geometry-guided loss]{
\begin{minipage}[t]{0.3\linewidth}
\centering
\includegraphics[width=\linewidth]{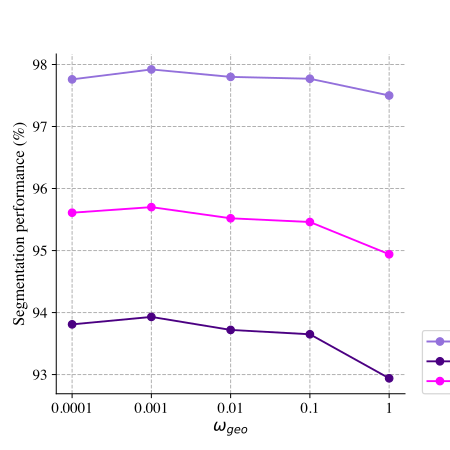}
\label{geo-guided}
\end{minipage}%
}%
\subfigure[Ratios of high-curvature points]{
\begin{minipage}[t]{0.3\linewidth}
\centering
\includegraphics[width=\linewidth]{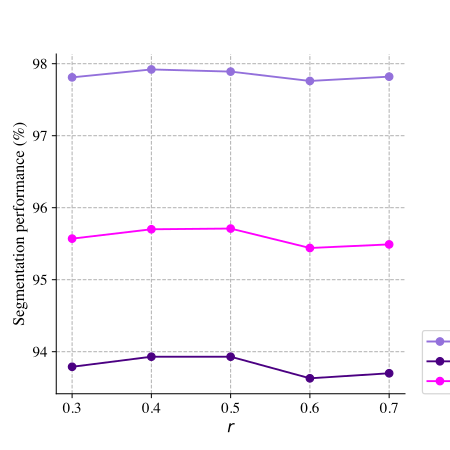}
\label{point ratio}
\end{minipage}%
}%
\subfigure[Auxiliary segmentation loss]{
\begin{minipage}[t]{0.3\linewidth}
\centering
\includegraphics[width=\linewidth]{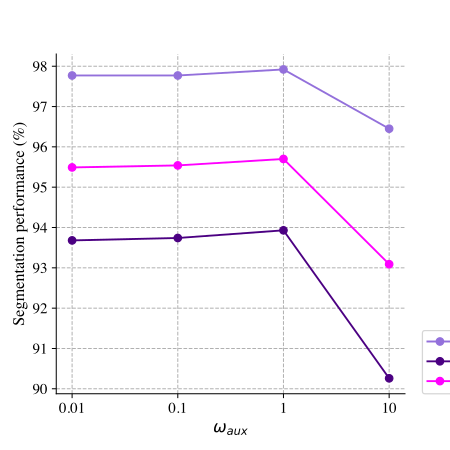}
\label{aux-branch weight}
\end{minipage}
}%
\centering
\caption{Ablation results for different hyperparamters. (a) Weights of the geometry guided loss. (b) Ratios of the points with high point curvature in the geometry guided loss. (c) Weights of the auxiliary segmentation loss. }
\end{figure*}

\section{Results and Discussion}
\label{sec 4}
\subsection{Dataset and Experimental Setup}  \label{setup}
\paragraph{\textbf{Dataset}}
Our IOS dataset is characterized by large scale and high resolution. Specifically, we collected 16,000 3D IOS mesh dental models during 2018-2021 in China, with evenly distributed maxillary and mandible scans manually labeled by human experts, each consisting of 100,000 to 350,000 triangular mesh faces. We randomly split the dataset into 12,000 scans for training, 2,000 for validation, and 2,000 for testing in which the number of maxillary and mandible jaws are not strictly equal. To the best of our knowledge, this is the largest IOS dataset for 3D tooth segmentation. 
Furthermore, we collect an external validation dataset with 200 cases to evaluate the clinical applicability of TFormer in real-world clinical situations.

\paragraph{\textbf{Network Architectures}}
For the point embedding module, we use two convolution layers with kernel size $1\times1$ and output dimension $d_0=64$ for $MLP_{pe}$. For the two \textit{EdgeConv} layers, we set  $k=32$ in the k-nearest neighbor, and the output dimensions are 64 and 128 ($d_e=128$), respectively. Besides, four self attention layers where $d_q=d_k=32$ and $d_v=128$ are employed in the encoder module. Both segmentation heads $MLP_{seg}$ and $MLP_{aux}$ are implemented with three convolution layers with kernel size $1\times1$ whose output dimensions are respectively 512, 256 and 33/2 (33 for $MLP_{seg}$ and 2 for $MLP_{aux}$). All convolution layers are all followed by batch normalization and ReLU activation function. We use dropout with a rate of 0.5 in $MLP_{seg}$ and $MLP_{aux}$.

\paragraph{\textbf{Training and Inference Settings}}
In the training stage, we first randomly sample 10,000 points in the raw tooth point cloud $\widetilde{P}$, i.e., $N=10,000$. We employ the stochastic gradient descent (SGD) optimizer with an initial learning rate of 0.1 that decays until 0.001 with the cosine annealing strategy during training. The network is trained for 200 epochs. 
In inference, we randomly partition all points in $\widetilde{P}$ into several batches each with $N=10,000$ points. If $\widetilde{N} \neq cN, where \; c \in \mathbb{N}_{+}$, we randomly select other points in $\widetilde{P}$ to make the above equations hold. We follow the literature to use the overall point-level classification accuracy (Acc), mean intersection over union (mIoU), and dice similarity coefficient (DSC) to evaluate the segmentation performance (\citet{meshsegnet,zhang2021tsgcnet,jdr}).

\subsection{Tooth Segmentation Performance}

We conducted comprehensive experiments by comparing our TFormer to seven representative and state-of-the-art baselines from three categories: 1) neural networks for point clouds; 2) transformers for point clouds; 3) domain-specific architectures for 3D tooth segmentation. As for traditional neural networks, we chose PointNet++ (\citet{pointnet++}) and the advanced DGCNN model (\citet{dgcnn}), as both of them are widely-used for lots of point cloud tasks. Moreover, we included the point transformer (\citet{zhao2021point}) and PVT (\citet{zhang2021pvt}) models as transformer-based baselines, which achieve state-of-the-art performance on various tasks. As for domain specific architectures, we included all the recently proposed deep learning based 3D tooth segmentation architectures whose source codes are publicly available, i.e., MeshSegNet (\citet{meshsegnet}), TSGCNet (\citet{zhang2021tsgcnet}) and DC-Net (\citet{jdr}). An interesting finding is that these architectures were evaluated in different settings in their original papers, but not systematically evaluated with large-scale dataset yet. All the baselines were implemented with their original source codes, and followed the same evaluation protocols in \ref{setup} for fair comparison. 


The quantitative segmentation results of TFormer and the baselines are presented in Table \ref{comparison with baseline}. 
We can make the following observations. First, TFormer obtained an overall accuracy of 97.97\%, mIoU of 94.34\%, Dice similarity coefficient of 96.01\%, outperforming existing best-performing point transformer model (\citet{zhao2021point}) by 0.12 \% in accuracy, 1.19 \% in mIoU and 0.93 \% in DSC. Such an improvement is surely significantly considering the complicated real-world cases in our large-scale dataset and the relatively high performance of point transformer with an mIoU of 93.30\%. Moreover, TFormer consistently surpassed all the baselines in terms on both mandible and maxillary in terms of all metrics, demonstrating its universal effectiveness. 

Another observation is mainly about different network architectures. We can notice that though MeshSegNet, TSGCNet, and DCNet are all domain-specific 3D tooth segmentation models, their performance, though on par with the PVT and DGCNN model, is worse than the point transformer model. This is also consistent with the superior performance of transformer-based models on standard point cloud processing tasks, which could be mainly attributed to the larger dataset and powerful attention mechanism that better captures global dependencies. Hence, though models like MeshSegNet adopt some task-specific designs to  achieve good performance, they still lag behind point transformer when a huge amount of data samples are available. In contrast, our TFormer employed the attention mechanism for point representation learning, and meanwhile, adopted task-specific architectures and geometry guided losses to further boost the performance. 

We also want to point out that MeshSegNet did not reach the performance claimed by \citet{meshsegnet}. There are two main reasons. First, we did not adopt any post-processing technique in our experiments a for fair comparison. Meanwhile, it's evaluated on such a large-scale dataset and requires the predictions of all mesh faces in a very high resolution rather than down-sampling in the original paper.

\begin{figure*}[!t]
\centering
\setlength{\abovecaptionskip}{0.cm}
\includegraphics[width=\linewidth]{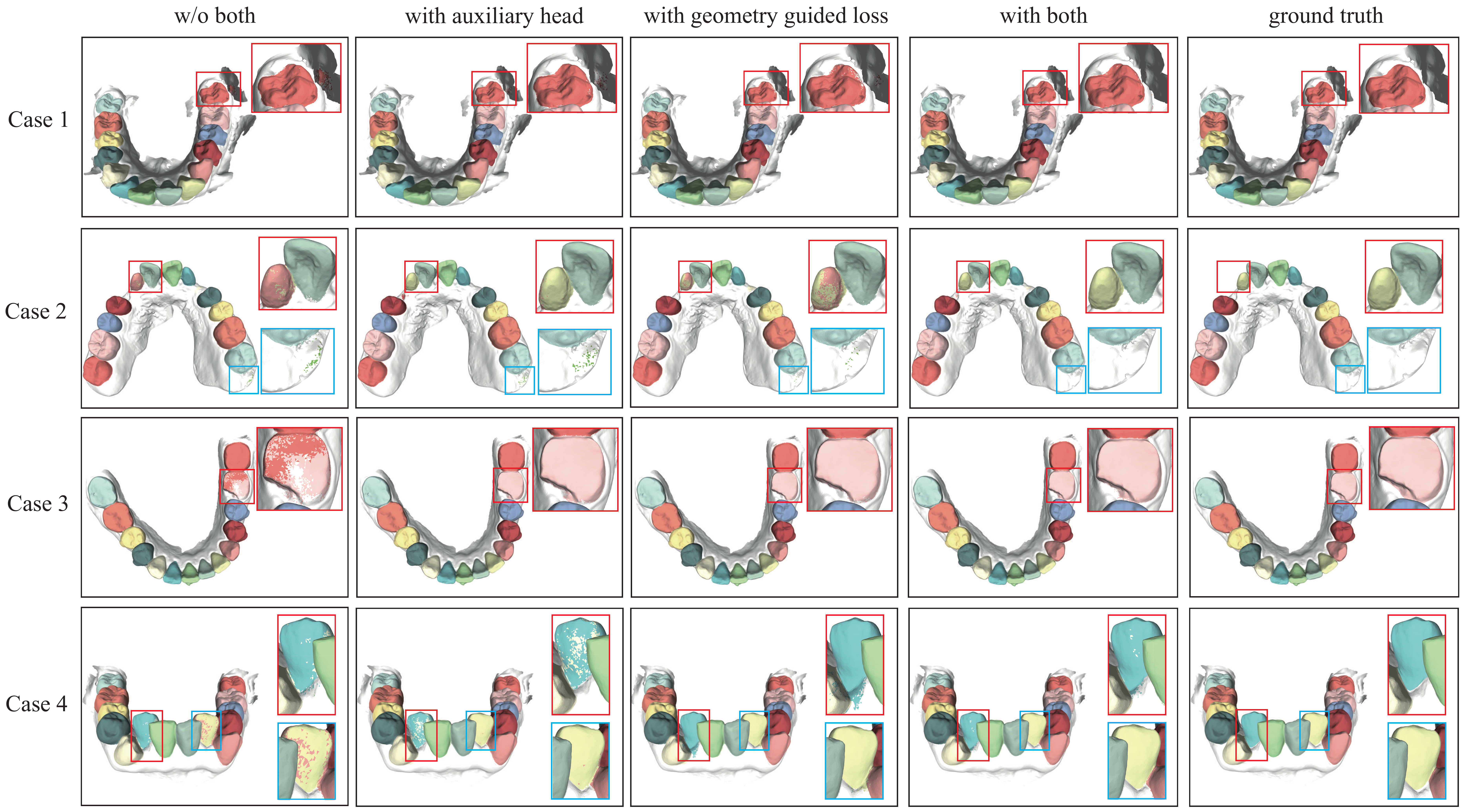}
\caption{Visualization results of TFormer with different main components.}
\label{vis base gingiva loss both gt}
\end{figure*}

\subsection{Clinical Applicability Test}

To demonstrate the effectiveness of TFormer in real-world scenarios, we conducted a clinical applicability test on an external validation dataset with 200 IOS scans. We generated the segmentation with five different models, which were evaluated by a committee of dentists with more than 5-year experience.  The results are shown in Table \ref{clinical analysis}. We can notice that TFormer significantly outperforms the other models regarding the clinical error rate, i.e., only 24\% of them are not clinically applicable. The feedback from dentists indicates that models such as TSGCNet cannot meet the requirement when dealing with the tooth-tooth and tooth-gingiva boundaries, while TFormer apparently handles them better. The point transformer and DCNet also showed promising  performance, but they are 
yet far behind our TFormer. 

As for the number of parameters and inference time, though TFormer has the second most parameters among all methods we tested, it is the second fastest method that only takes around 23 seconds to complete end-to-end inference, which is certainly acceptable in real-world clinical scenarios.

\subsection{Ablation Studies}

\subsubsection{Effectiveness of Geometry Guided Loss}
As reported in Table \ref{ablation-geo&aux}, introducing the geometric guided loss can improve the performance under all three metrics, e.g., the mIoU gets around a 0.4\% improvement. These improvements are also consistent with the better visualizations as illustrated in Fig. \ref{vis base gingiva loss both gt}. We can notice that the isolated mispredictions in the figures of the third column are greatly corrected compared to the first column, indicating the significance of perceiving the boundary regions with this loss. 

Furthermore, we demonstrated the universal effectiveness of the geometry guided loss for other architectures by adding it to DC-Net (\citet{jdr}). As shown in Table \ref{DCNet+loss}, the segmentation performance of DC-Net is also enhanced by 1.43\% in mIoU with this additional loss.

We conducted an ablation study on the weights of geometry guided loss, as reported in Fig.\ref{geo-guided}.  Notably, all Acc, mIoU, and DSC have similar trends as $\omega_{geo}$ varies. And the highest performance is achieved under all three metrics when $\omega_{geo} = 0.001$, where the model can obtain an Acc of 97.92\%, mIoU of 93.93\%, and DSC of 95.70\%. In addition, we summarize the result of using different ratios of high-curvature in Fig.\ref{point ratio}. The line chart shows that TFormer achieves the best performance when $r=0.4$. Overall, the performance is not very sensitive to the changes of both $\omega_{geo}$ and $r$ in a certain range.



\subsubsection{Effectiveness of the Auxiliary Segmentation Head}
The auxiliary segmentation head is designed to rectify the inaccuracy brought by mislabeling teeth and gingiva near their boundaries. As shown in Table \ref{ablation-geo&aux}, adding a loss for the auxiliary branch leads to about 0.4\% mIoU performance improvement. Additionally, Fig. \ref{vis base gingiva loss both gt} demonstrates that adding the auxiliary head to our network, i.e., the figures of the fourth column, can help refine the teeth-gingiva separation boundary and further fix prediction errors with only geometry guided loss, i.e., the third column.  
We also report the result of changing the weight of loss from the auxiliary segmentation head, as shown in Fig.\ref{aux-branch weight}. TFormer achieves the best performance when $\omega_{aux}=1$, and degrades a bit when $\omega_{aux}$ reduces to 0.1 or smaller. 


\begin{table*}[htb]
\centering
\setlength{\abovecaptionskip}{0pt}
\setlength{\belowcaptionskip}{8pt}
\caption{Segmentation performance of TFormer under different training set scales.}
\label{training set scale}
\resizebox{0.7 \textwidth}{!}{
\begin{tabular}{c|ccc|ccc|ccc}
\hline
\multirow{2}{*}{Training set scale} &
  \multicolumn{3}{c|}{Mandible} &
  \multicolumn{3}{c|}{Maxillary} &
  \multicolumn{3}{c}{All} \\ \cline{2-10} 
        & mIoU & DSC  & Acc  & mIoU & DSC  & Acc   & mIoU & DSC  & Acc \\ \hline \hline
500     &86.36 &89.74 &95.45 &89.22 &91.69 &96.57  &87.86 &90.76 &96.04 \\
1,000   &90.60 &93.13 &96.72 &92.84 &94.77 &97.50  &91.78 &93.99 &97.13 \\
2,000   &92.15 &94.28 &97.31 &94.05 &95.79 &97.92  &93.15 &95.08 &97.63 \\
4,000   &93.10 &95.05 &97.69 &94.68 &96.28 &98.12  &93.93 &95.70 &97.92 \\
8,000   &93.27 &95.15 &97.67 &94.83 &96.40 &98.16  &94.09 &95.81 &97.92 \\
12,000  &\textbf{93.53} &\textbf{95.36} &\textbf{97.72}  &\textbf{95.07} &\textbf{96.60} &\textbf{98.20}  &\textbf{94.34} &\textbf{96.01} &\textbf{97.97}\\ \hline
\end{tabular}
}
\end{table*}

\begin{figure*}[!htb]
\centering
\setlength{\abovecaptionskip}{0.cm}
\includegraphics[width=\linewidth]{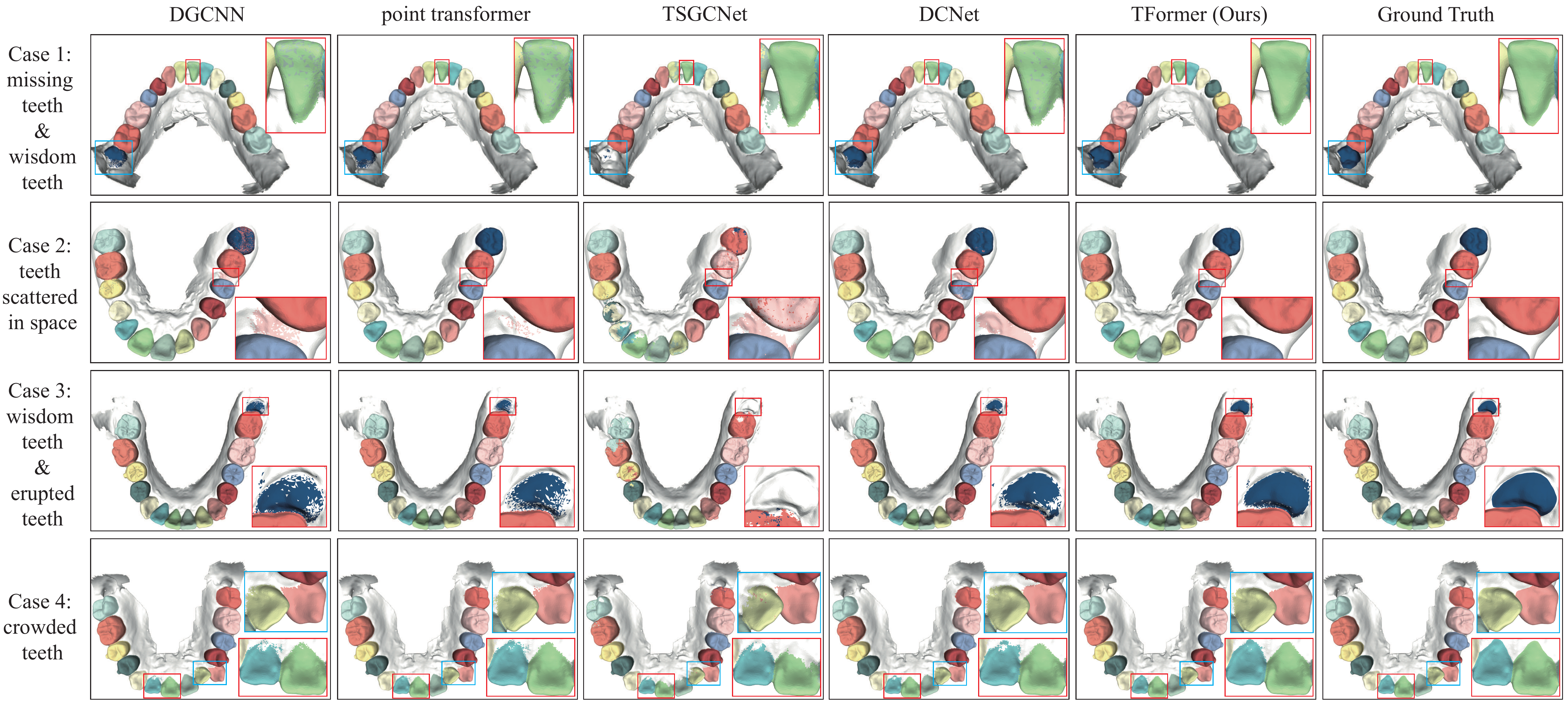}
\caption{Visualization of the dental model segmentation results with different methods.}
\label{vis of different methods}
\end{figure*}

\begin{figure*}[!htb]
\centering
\setlength{\abovecaptionskip}{0.cm}
\includegraphics[width=\linewidth]{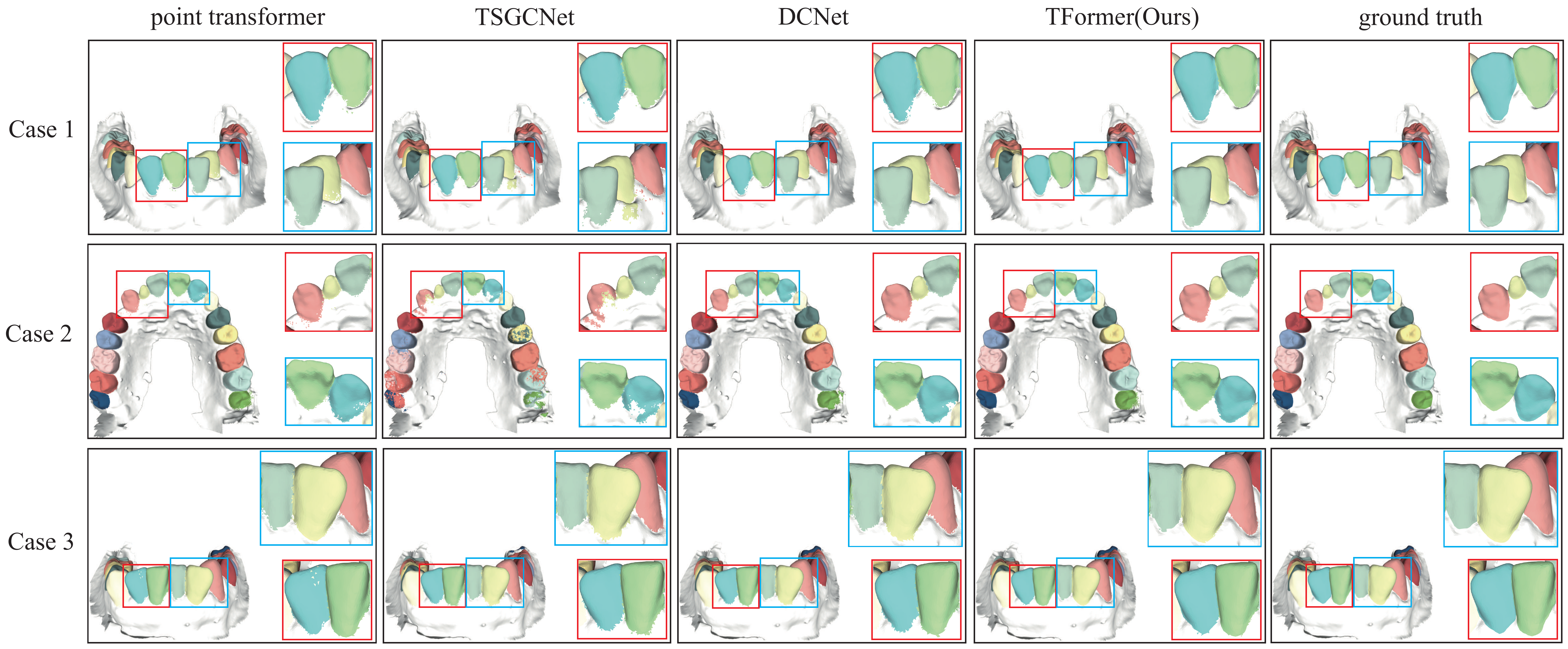}
\caption{Visualizations of boundary regions for different models. TFormer produces more smoothing boundary segmentation. }
\label{vis boundary cases}
\end{figure*}


\subsubsection{Effectiveness of Point Embedding}
The result of adopting different point embedding strategies is summarized in Table \ref{point-embed}. Apparently, embedding with purely MLP-based structures gives the worst result, 
since they are not able to capture sufficient local contextual information. On the other hand, with the EdgeConv module, the segmentation performance is greatly improved for all three metrics. Among them, the mIoU and DSC gain about 2\% performance improvement. Besides, as mentioned earlier, we find that combining MLP-based point embedding with EdgeConv can further enhance the performance. 

\subsubsection{Effectiveness on Different Training Set Scale}
In real-world orthodontic applications, large-scale training data may not be directly accessible due to privacy concerns. Therefore, to demonstrate our model's data efficiency, we train our model on datasets with different sizes, as shown in Table \ref{training set scale}. With only 500 training samples, TFormer is able to reach an overall accuracy of 96.04\%, mIoU of 87.86\%, DSC of 90.76\%, surpassing PointNet++, MeshSegNet and TSGCNet trained on 12,000 samples. Furthermore, TFormer trained with only 2,000 samples can almost outperform all previous models trained on 12,000 samples. Overall, these results demonstrate the exceptional data efficiency of our TFormer.

\begin{figure}[H]
\centering
\setlength{\abovecaptionskip}{0.cm}
\includegraphics[width=\linewidth]{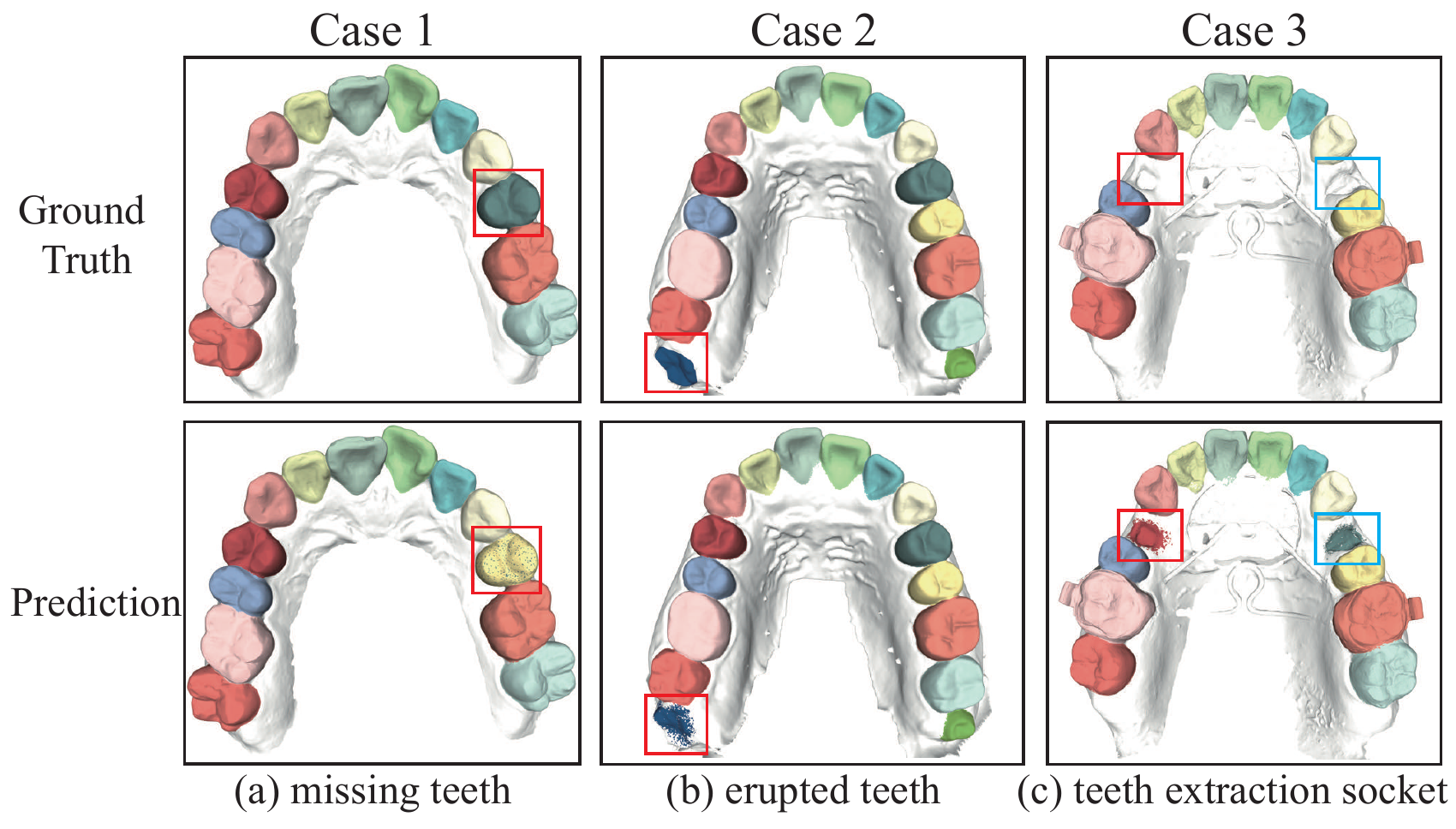}
\caption{Limitations of our model.}
\label{vis_limit}
\end{figure}

\subsection{Visualization}
\label{visualization}
We visualize the segmentation results and demonstrate the superiority of our method on various complicated dental diseases, as illustrated in Fig. \ref{vis of different methods}. The visualized diseases include the missing teeth, third-molars, dentural diastema, erupted teeth, crowded teeth etc., which usually lead to complicated dentitions and cause difficulties for segmentation in clinical applications. The baseline models unavoidably produce false or incomplete predictions or even fail to identify an entire third-molar, while TFormer can yield more accurate segmentation results and smoother boundaries, corroborating great potential for clinical applications.

\begin{table}[H]
\centering
\setlength{\abovecaptionskip}{0pt}
\setlength{\belowcaptionskip}{8pt}
\caption{Statistical analysis results of IoU(\%) for all 32 tooth categories.}
\resizebox{\linewidth}{!}{
\begin{tabular}{c|cccccccc}
\hline
Tooth Label              & 11    & 12    & 13    & 14    & 15    & 16    & 17    & 18    \\
IoU (\%)               & 96.01 & 94.59 & 94.95 & 94.40 & 94.74 & 95.39 & 94.76 & 92.46 \\

\hline \hline

Tooth Label              & 21    & 22    & 23    & 24    & 25    & 26    & 27    & 28    \\
IoU (\%)               & 96.19 & 95.26 & 95.68 & 94.34 & 94.94 & 96.37 & 95.99 & 93.59 \\

\hline \hline

Tooth Label              & 31    & 32    & 33    & 34    & 35    & 36    & 37    & 38    \\
IoU (\%)               & 90.79 & 92.85 & 96.01 & 94.33 & 94.14 & 95.19 & 94.03 & 91.94 \\

\hline \hline

Tooth Label              & 41    & 42    & 43    & 44    & 45    & 46    & 47    & 48    \\
IoU (\%)               & 90.24 & 91.32 & 95.14 & 93.42 & 93.55 & 94.45 & 93.86 & 91.67 \\

\hline
\end{tabular}
}
\label{Statistical analysis}
\end{table}

\subsection{Discussion}

A marked contribution of TFormer is the superior ability to delineate the complicated tooth-tooth and tooth-gingiva boundaries across patients with different diseases, which is empowered by the transformer-based architecture, auxiliary segmentation head, and geometry guided loss. Besides the aforementioned results such as a clinical applicability test, we further verify such property by randomly selecting three test samples and visualizing the segmentation from different models, as illustrated in Fig. \ref{vis boundary cases}. We can notice that TFormer can better recognize the boundaries and produce much smoother boundaries without jaggy segmentation as in baseline models. 


Nevertheless, there remain some limitations. First, our model would commit some mistakes in some hard cases, as illustrated  in Fig. \ref{vis_limit}. For example, our model recognizes the first premolar as the second premolar due to the confusing tooth arrangement incurred by the missing second premolar (the red boxes in Case 1). Additionally, it also fails to discover the erupted wisdom tooth in the left side (the red boxes in Case 2)  completely and accurately enough, and misjudges the sunken gingiva and alveolar bone regions or teeth extraction socket as the teeth (the red and blue boxes in Case 3). 

Moreover, we conduct a performance statistical analysis for 32 tooth categories and the corresponding IoUs are reported in Table \ref{Statistical analysis}. It is clear that our model has good segmentation performance on most tooth categories, but relatively poor performance for teeth with FDI numbers 31, 32, 41, 42, 38, and 48, i.e, the mandible incisors and third molars. The relatively worse performance on the third molars is mainly attributed to the large missing rate, i.e., more than 60\% of the patients do not have third molars. Meanwhile, the incisors in the mandible would easily exhibit severe dentition crowding or missing teeth, resulting in worse segmentation performance. Future work which explores  the distribution gap among different dental diseases is of high necessity to further boost the performance. 


\section{Conclusion}
\label{sec 5}
In this paper, we propose an automated Transformer-based model, named TFormer, towards more clinically applicable tooth segmentation on large-scale and high-resolution 3D IOS data. With the point embedding module and attention mechanism, TFormer can effectively capture both local and global features to distinguish teeth and gingiva in complicated real-world cases. Besides, we propose a novel geometry guided loss based on curvature information to handle the error-prone boundary regions, and devise multi-task segmentation heads to further improve the boundary recognition accuracy. The method is evaluated on a large-scale dataset with 16,000 IOS scans, the largest IOS dataset to the best of our knowledge. Comprehensive experiments and clinical applicability tests show that TFormer obtains state-of-the-art performance and exhibits superior capability in real-world clinical applications. Our work corroborates the great potential of cutting-edge deep learning techniques in future digital dentistry. \\

\bibliographystyle{model2-names.bst}

\bibliography{ref}

\end{multicols}



\end{document}